\documentclass[conference]{IEEEtran}
\IEEEoverridecommandlockouts
\usepackage{cite}
\usepackage{amsmath,amssymb,amsfonts}
\usepackage{algorithmic}
\usepackage{graphicx}
\usepackage{textcomp}
\usepackage{xcolor}
\usepackage{comment}
\usepackage{tikz}
\usepackage{pgfplots}
\usepackage{textgreek}
\pgfplotsset{compat=1.17}
\usetikzlibrary{patterns}
\usepackage{censor}

\usepackage{enumitem}
\usepackage[most]{tcolorbox}
\usetikzlibrary{shadows.blur}
\usetikzlibrary{calc}
\usepackage[font=small,labelfont=bf]{caption}

\tcbset{
    mymsg/.style={
        enhanced,
        colframe=white, frame hidden,
        boxrule=0pt,
        arc=18pt,
        outer arc=18pt,
        boxsep=2pt, left=8pt, right=8pt, top=6pt,
        colback=#1!25,
        fuzzy shadow={1mm}{-0.7mm}{0mm}{0.1mm}{black!40!white},
        width=0.9\linewidth,
        before skip=-0.5mm, 
    },
    expertstyle/.style={mymsg=green, enlarge left by=10mm},
    llmstyle/.style={mymsg=blue, enlarge right by=10mm}
}

\pgfplotsset{compat=1.17}

\usepackage{url}
\usepackage{fancyhdr}

\def\BibTeX{{\rm B\kern-.05em{\sc i\kern-.025em b}\kern-.08em
    T\kern-.1667em\lower.7ex\hbox{E}\kern-.125emX}}
\begin{document}

\title{LLMs as Agentic Cooperative Players in Multiplayer UNO}

\author{\IEEEauthorblockN{Yago Romano Martinez}
\IEEEauthorblockA{\textit{Department of Computer Science} \\
\textit{Tennessee Tech University}\\
Cookeville, U.S. \\
yromanoma42@tntech.edu}
\and
\IEEEauthorblockN{Jesse Roberts}
\IEEEauthorblockA{\textit{Department of Computer Science} \\
\textit{Tennessee Tech University}\\
Cookeville, U.S. \\
jtroberts@tntech.edu}
}

\maketitle

\begin{abstract}

LLMs promise to assist humans, not just by answering questions, but by taking direct, agentic action to support a desired outcome. However, this has not been investigated in strategic settings with the LLM acting as a non-player character (NPC) in support of the player's goal. We ask, can a large language model based agent help another accomplish their strategic goal as an active participant? We test this question by engaging an LLM in UNO, a turn‑based card game, asking it not to win but instead help another player to do so. We develop a tool that allows decoder‑only LLMs to participate as agents within the RLCard game environment. We evaluate models ranging from small (1B parameters) to large (70B parameters) and explore how model scale impacts performance. We find that while all models were able to successfully outperform a random baseline when playing UNO, few were able to significantly aid another player. 

\end{abstract}

\section{Introduction}

The rapid advancement of large language models (LLMs) has revolutionized artificial intelligence applications, enabling machines to generate human‑like text~\cite{radford2018improving,radford2019language,mann2020language}, reason through complex tasks~\cite{wei2022chain}, and interact dynamically with users. Among these models, decoder‑only architectures \cite{roberts2024powerful} such as the GPT family~\cite{radford2018improving,brown2020language} have demonstrated exceptional performance in sequential decision‑making tasks due to their ability to predict next‑token outputs based on contextual understanding. While LLMs have been widely studied in language generation and autonomous problem solving~\cite{chowdhery2023palm}, their role as agentic players in cooperative gameplay with teammates remains underexplored. 

This project begins to address that gap and provide a tool for others to build upon. We extend RLCard UNO \cite{zha2019rlcard} to allow decoder‑only LLMs play turn-by-turn as an autonomous agent\footnote{LLM UNO: \url{https://github.com/yagoromano/llm-uno} \\ Reproducibility: \url{https://github.com/yagoromano/llm-uno-experiments} }. This framework allows any LLM to function as a complete UNO player, receiving the full game state and selecting actions via text prompts. With this tool, we investigate whether LLMs can not only play the game independently, but also act as cooperative partners to assist another player in reaching their goal.

We study two settings: (1) autonomous play, where the LLM competes head-to-head against a random agent to assess baseline competence, and (2) cooperative play, where the LLM is paired with a rule-based teammate in a three-player game to evaluate whether it can improve its partner’s win rate. Across both tasks, we explore how model scale (1B to 70B parameters) and prompting technique (cloze vs. counterfactual) influence the LLM’s performance.

This work centers around four core research questions:
\begin{enumerate}[%
  label=\textbf{RQ\arabic*:},%
  labelindent=1em,%
  leftmargin=*%
]
  \item Can a pretrained decoder‑only LLM, when playing autonomously, achieve a higher win rate than a random agent in UNO?
  \item Can the same LLM, when deployed as a cooperative partner, provide statistically significant strategic support that improves a teammate’s performance?
  \item To what extent does scaling the LLM’s size impact its autonomous and cooperative effectiveness?
  \item How do prompting techniques—specifically cloze prompting and counterfactual prompting—affect the quality of the LLM’s decision‑making in both scenarios?
\end{enumerate}

By embedding LLMs in a structured multi-agent game, this work moves beyond passive language understanding and toward active collaboration—laying the groundwork for future LLMs to serve as dynamic, real-time partners in complex, goal-driven environments.

\section{Background}

\subsection{Introduction to Large Language Models (LLMs)}

Large language models (LLMs) are sophisticated systems that generate human-like text by learning patterns from massive datasets. They have significantly influenced various applications, such as language translation, question answering, automated summarization, and interactive decision-making. Among various architectures, decoder-only LLMs stand out due to their simplicity, scalability, and demonstrated effectiveness in sequential tasks, particularly text generation.

\subsection{Architectures of Large Language Models: An Overview}

LLM architectures typically fall into three distinct categories:

\begin{enumerate}
    \item \textbf{Encoder-only models} (e.g., BERT~\cite{devlin2019bert}): Focus on understanding input text without directly generating sequences.
    \item \textbf{Encoder-decoder models} (e.g., T5~\cite{raffel2020exploring}): Handle input-to-output transformations such as language translation.
    \item \textbf{Decoder-only models} (e.g., GPT family~\cite{radford2018improving,brown2020language}): Generate text by sequentially predicting the next word based on previous words.
\end{enumerate}

While all three architectures have proven successful across various tasks, only the encoder-decoder and decoder-only architectures are suitable for natural language generation \cite{roberts2024powerful}. This project focuses specifically on \textbf{decoder-only models} due to their prevalence and existing work on their inclusion in game settings \cite{roberts-etal-2025-large}. Further, these models are particularly well-suited for sequential environments like UNO, where actions unfold one at a time and each move depends on previous game context.

\subsection{How Decoder-only Models Work}

To understand how these models function in practice, we examine the key components of a typical decoder-only transformer architecture. These models build on the original transformer design introduced by Vaswani et al.~\cite{vaswani2017attention}, using masked self-attention to predict the next token in a sequence based solely on prior context.

Key components:
\begin{itemize}
    \item \textbf{Tokenization \& Embeddings}: The text is split into smaller pieces called subwords (using Byte‑Pair Encoding~\cite{sennrich2015neural}), each converted into a numeric vector and tagged with its position so the model knows word order.
    \item \textbf{Self‑Attention \& Feed‑Forward Layers}: In each transformer layer, every token “looks” at all other tokens to decide which are most important (self‑attention), then a simple neural network (feed‑forward) updates each token’s vector.
    \item \textbf{Autoregressive Decoding}: When generating text, the model predicts one token at a time—choosing the next token from a probability distribution based on the tokens it has already produced.
\end{itemize}

\subsection{RLCard as an Experimental Platform}

RLCard~\cite{zha2019rlcard} is an open-source platform specifically designed for reinforcement learning research in card-game environments. It provides a standardized framework to implement, evaluate, and benchmark different AI strategies, making it an ideal environment to test the capabilities of decoder-only LLM agents.

RLCard’s primary functionalities include:

\begin{itemize}
    \item \textbf{Game State Representation:} RLCard defines a clear representation of game states, actions available to players, and associated game rules. It simplifies the process of integrating agents by providing consistent game-state formats.
    
    \item \textbf{Agent Interfaces:} Standardizes the way agents interact with the game, providing consistent methods for observing states and performing actions. It also includes built-in random and rule-based agents, which we use as baselines in our turn-order and LLM performance experiments.
    
    \item \textbf{Performance Metrics:} RLCard provides evaluation metrics such as win rates, enabling consistent and objective comparisons across various agents and games.
\end{itemize}

This standardized setup makes RLCard particularly useful for evaluating how well LLMs perform both individually and cooperatively in structured multiplayer games like UNO.

\begin{figure}
  \centering
  \begin{tikzpicture}
    \begin{axis}[
      width=\linewidth,                             
      title={\large First‑Player Advantage in Three‑Player UNO},
      title style={
        align=center,
        font=\small,
        text width=10cm
      },
      ybar,
      bar width=12pt,
      symbolic x coords={Rand,Rule},
      xtick=data,
      ylabel={Win Rate (\%)},
      ymin=30, ymax=36,
      enlarge x limits=0.3,
      nodes near coords,
      legend style={at={(rel axis cs:0.5,0.85)},anchor=south,legend columns=3}
    ]
      \addplot+[ybar,bar shift=-18pt] coordinates {
        (Rand,34.58) (Rule,34.38)
      };
      \addplot+[ybar,bar shift=0pt] coordinates {
        (Rand,33.33) (Rule,33.69)
      };
      \addplot+[ybar,bar shift=18pt] coordinates {
        (Rand,32.09) (Rule,31.93)
      };
      \legend{Player0,Player1,Player2}
    \end{axis}
  \end{tikzpicture}
  \caption{Win rates by seat for three‑player random agent and rule‑based agent matches (10,000 games each), highlighting the first-player advantage in UNO.}
  \label{fig:3player_baseline}
\end{figure}
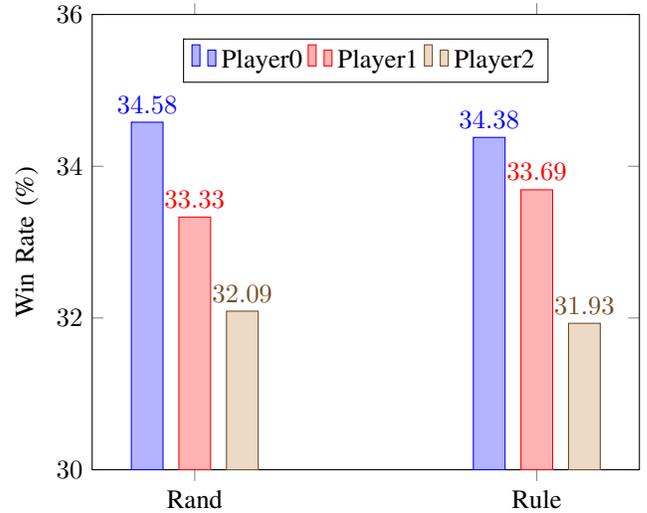

\begin{figure}
  \centering
  \begin{tikzpicture}
    \begin{axis}[
      width=\linewidth,
      title={\large First‑Player Advantage in 1v1 UNO},
      title style={
        align=center,
        font=\small,
        text width=10cm
      },
      ybar,
      bar width=12pt,
      symbolic x coords={Rule-vs-Rule,Rand-vs-Rand},
      xtick=data,
      ylabel={Win Rate (\%)},
      ymin=48, ymax=52,
      nodes near coords,
      enlarge x limits=0.3,
      legend style={at={(rel axis cs:0.5,0.85)},anchor=south,legend columns=2}
    ]
      \addplot+[ybar,bar shift=-10pt] coordinates {(Rule-vs-Rule,50.58) (Rand-vs-Rand,51.04)};
      \addplot+[ybar,bar shift=+10pt] coordinates {(Rule-vs-Rule,49.42) (Rand-vs-Rand,48.96)};
      \legend{Player 1, Player 2}
    \end{axis}
  \end{tikzpicture}
  \caption{Win rates by seat in 1v1 UNO matches (10,000 games each), showing a consistent first-player advantage for both Rule-Based and Random agents.}
  \label{fig:head_to_head}
\end{figure}
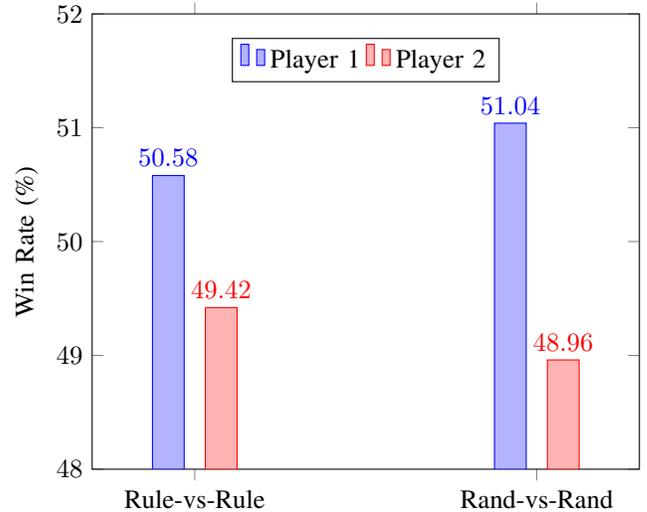

\subsection{UNO and Opening-Player Advantage}

UNO is a popular multiplayer turn‑based card game known for its simple rules and quick gameplay. Players compete to be the first to discard all their cards, using a combination of numbered cards and special action cards—such as Skip, Reverse, and Draw Two—to influence turn order and pacing. Though we hope to study the efficacy of LLMs in more complex, real-time games, UNO's turn-based nature and simple mechanics makes it an ideal environment for evaluating early self-centric and cooperative agentic LLM success. 

Despite these benefits, the player order is heavily influential in UNO, providing an advantage to the opening player. Our baseline experiments reveal that the opening player wins most often when controlled for type of agent. To provide fair evaluation, we account for this bias so our results show real success, not simply order-based benefit.

To quantify the effect, we used RLCard’s built‑in random and rule‑based agents to play 10,000 baseline games in both three‑player and head‑to‑head matchups. Figures\ref{fig:3player_baseline} and \ref{fig:head_to_head} display the win‑rate distributions for these scenarios. The results confirm that random rule‑based agents are impacted by the first‑turn benefit. We adjust all subsequent LLM analyses to report performance relative to fixed player positions.

\section{Related Works}


Here we survey a number of papers that have investigated related questions involving LLMs in gameplay. However, the present paper is the first to explicitly study LLMs tasked with supporting another player in a competitive environment.

Xu et al.~\cite{xu2025agents} introduced the PORTAL framework, a notable approach to integrating decoder-only large language models (LLMs) into gaming. PORTAL automatically generates behavior trees for AI agents across a variety of 3D video game environments using decoder-only architectures. By framing complex strategic decision-making as a language modeling task, PORTAL demonstrates both the adaptability and generalization capabilities of LLM-based agents, while also improving interpretability and rapid policy deployment.

In collaborative gameplay contexts involving human players, Rao et al.~\cite{rao2024collaborative} implemented GPT-4-driven non-player characters (NPCs) within the sandbox game Minecraft. Their work demonstrated how collaborative behaviors naturally emerge from human interactions with language-driven agents, highlighting the benefits and viability of language-based cooperation. The study also highlighted key challenges in integrating non-linguistic inputs into the agent’s understanding.

Additionally, Buongiorno et al.~\cite{buongiorno2024pangea} demonstrated the success of decoder-only LLMs for dynamic narrative generation in turn-based role-playing games with their PANGeA system. PANGeA generates contextually coherent narrative elements and NPC interactions, adapting to varied player inputs while preserving narrative consistency throughout gameplay.

While each of these studies demonstrates significant advancements in integrating decoder-only LLMs within gaming contexts, they do not yet fully address the potential of LLMs to support explicit, structured cooperative strategies directly alongside human players. Specifically, PORTAL primarily addresses autonomous agent behaviors without incorporating human-agent cooperation. Rao et al.'s~\cite{rao2024collaborative} work in Minecraft focuses mainly on open-ended interactions rather than structured cooperative objectives. Likewise, PANGeA primarily supports narrative-driven interactions rather than explicit strategic collaboration.

In contrast, our work directly examines how decoder-only LLMs can provide strategic assistance to human players within a controlled cooperative gaming environment. Using the RLCard UNO framework, we evaluate whether LLM-generated strategic guidance can improve human player performance and goal achievement compared to random or baseline strategies, addressing a critical gap in existing research.

\section{Methodology}

This section outlines the approach taken to evaluate the performance of LLMs acting as players within UNO. To explore the impact of model scale and architecture on decision quality, we selected four open‑source decoder-only LLMs: LLaMa3.2-1B (1 billion parameters) as a compact baseline, LLaMA3.1‑8B (8 billion parameters), Mistral-Small-24B-Instruct-2501 (24 billion parameters), and LLaMA3.3‑70B-Instruct (70 billion parameters) as a larger‑scale reference. Specifically, we evaluate two main scenarios: the LLM playing independently against a random agent, and the LLM acting as a cooperative agent to play alongside and assist a teammate. In both cases, we aim to assess whether the LLM can make effective decisions that improve gameplay outcomes. This section describes the prompting strategies, the experimental setup for both scenarios, the system architecture and execution modes (single-node and multi-node), the integration of the LLM into the RLCard UNO framework, and the statistical methods used to evaluate its performance.

\subsection{Prompting Methods}

To guide the LLM’s decision-making process during gameplay, we developed structured prompts that comprised three parts. First, an explicit statement of the LLM’s task and role as an UNO player charged with assisting a specific rule-based agent to victory. Second, a concise summary of UNO rules and the effects of special action cards, including the overall win condition. Third, the current game state data—number of players, last played card, hand contents, next player, recent moves, and legal actions. Finally, the LLM was asked to choose the best action according to the specified prompting method.  

The game state information was extracted from RLCard and reformatted for readability. While RLCard encodes cards using shorthand (e.g., “r-5” for red 5), we expanded these into full descriptions to improve the model’s comprehension. An example of the complete prompt format is shown in Figure~\ref{fig:LLM_interaction}.

\begin{figure}[h!]
    \centering
    \begin{tcolorbox}[expertstyle]
        \hfill \underline{\textbf{\large Game State Prompt Data Sample}} \\
        \textbf{Current Game State}\\
        \textbf{Number of Players:} \\
        3 (Player 0, Player 1, Player 2) \\
        \textbf{Number of Cards per Player:} \\
        Player 0: 3, Player 1: 5, Player 2: 7  \\
        \textbf{Last Played Card:} \\
        green skip (played by Player 1) \\
        \textbf{Your Hand:} \\
        red 1, green 2, yellow 1, green 9, yellow 7, \\
        red skip, yellow 5 \\
        \textbf{Next Player:} Player 1 \\
        \textbf{Recent Moves (last 5 cards played):} \\
        Player 0: draw \\
        Player 2: yellow 3 \\
        Player 1: green wild \\
        Player 0: green draw\_2 \\
        Player 1: green skip  \\
        \textbf{Legal Actions:} A: green 2, B: green 9, C: red skip
    \end{tcolorbox}

    \vspace{2mm}

    \begin{tcolorbox}[llmstyle]
        \underline{\textbf{\large LLM Action Selection and Probabilities}} \\
          {[}Shift 0{]} Letter A: 0.1103 → action 'g-2'\\
          {[}Shift 0{]} Letter B: 0.0758 → action 'g-9'\\
          {[}Shift 0{]} Letter C: 0.6345 → action 'r-skip'\\
          \textbf{-------------------------------------------------}\\
          {[}Shift 1{]} Letter A: 0.0609 → action 'g-9'\\
          {[}Shift 1{]} Letter B: 0.5780 → action 'r-skip'\\
          {[}Shift 1{]} Letter C: 0.1877 → action 'g-2'\\
          \textbf{-------------------------------------------------}\\
          {[}Shift 2{]} Letter A: 0.3065 → action 'r-skip'\\
          {[}Shift 2{]} Letter B: 0.3936 → action 'g-2'\\
          {[}Shift 2{]} Letter C: 0.1278 → action 'g-9'\\
        \textbf{=== CUMULATIVE SCORES ===}\\
          g-2: 0.6915\\
          g-9: 0.2645\\
          r-skip: 1.5191\\
          → Selected: r-skip
    \end{tcolorbox}

    \caption{Sample decision trace from Mistral‑24B in a UNO match: the model cycles through legal actions (A: green 2, B: green 9, C: red skip) over three prompt shifts, reports per-shift probabilities, aggregates cumulative scores, and selects the highest‑scoring move.}

    \label{fig:LLM_interaction}
\end{figure}

To drive the model’s action selection, we applied two prompting strategies inspired by Moore et al.~\cite{moore2024reasoning}: cloze prompting and counterfactual prompting. These methods determine how the model interprets the prompt and evaluates its legal actions during gameplay.

\textbf{Cloze Prompting:} In this method, legal actions were labeled with sequential letters (A, B, C, etc.), and the LLM was instructed to choose the letter corresponding to the best move. Only one token was allowed in the output, and the highest-probability token from the set of allowable actions was selected as the action. However, the tested models exhibited a consistent bias toward the first listed option—a known issue referred to as Base Rate Probability (BRP)\cite{moore-etal-2024-base}. BRP is the intrinsic likelihood of a token being selected based on minimal context, regardless of the surrounding prompt. Tokens like “A” or “yes” tend to have higher BRP than others, which can skew results even when no semantic preference exists. To reduce this bias, we implemented a rotation mechanism: the prompt was repeated as many times as there were legal actions, each time with a shifted ordering of the actions and a different letter assigned to each one. The cumulative token probability across these permutations was then used to select the final move (see Figure~\ref{fig:LLM_interaction}, LLM Action Selection and Probabilities).

\textit{For example, suppose the legal actions are:} 
\texttt{\textcolor{red}{red 2}}, \texttt{\textcolor{red}{red 5}}, and \texttt{\textcolor{blue}{blue 3}}.  
The prompt is repeated with the legal actions rotated in the following way:
\begin{itemize}
    \item \texttt{A: \textcolor{red}{red 2}, B: \textcolor{red}{red 5}, C: \textcolor{blue}{blue 3}}
    \item \texttt{A: \textcolor{red}{red 5}, B: \textcolor{blue}{blue 3}, C: \textcolor{red}{red 2}}
    \item \texttt{A: \textcolor{blue}{blue 3}, B: \textcolor{red}{red 2}, C: \textcolor{red}{red 5}}
\end{itemize}

In each case, the model distributes probabilities to every legal action, and we record the probabilities assigned to each one. Summing the probabilities across all rotations allows for a more balanced comparison that accounts for BRP-induced bias.

\textbf{Counterfactual Prompting:} Similar to cloze prompting, we queried the LLM as many times as there were legal actions. However, in this approach, each prompt asked specifically about one action and whether the model considered it a “good” or “bad” move. We then computed the difference in token probabilities (“good” minus “bad”) for each action, and selected the one with the highest average differential. To minimize BRP-related effects, no labels were used; instead, the model was prompted using full natural-language descriptions of each card.

We also evaluated extended prompt formats—such as few-shot examples and free-form paragraph descriptions of the game state—within both the cloze and counterfactual methods. In every case, the concise rule-style prompt produced the highest win rates, so we retained that format for all experiments.

\subsection{Integration with RLCard}

To embed the LLM as a player in UNO, we extended the RLCard UNO framework by patching its game class to support arbitrary player counts and updating its payoff logic. Our patched game class overrides the payoff logic to correctly instantiate all players and assign rewards appropriately, instead of the default two‑player payoff behavior. We also implemented an agent subclass conforming to RLCard’s agent interface. On each call to the agent:
\begin{enumerate}
  \item Retrieves the RLCard observation dictionary, which includes each player’s hand, the last played card, the next player ID, the set of legal actions, and a short history of recent plays.
  \item Transforms this information into a concise natural‑language prompt (see Figure~\ref{fig:LLM_interaction}).
  \item Loads the pretrained LLM, if not already loaded, and performs inference to obtain logits (unnormalized scores for each possible next token) for the first output token.
  \item Applies a softmax to those logits and uses greedy decoding to select the highest‑probability action label (A, B, C, …) according to the prompting method.
  \item Maps the chosen label back to RLCard’s integer action index and returns it.
\end{enumerate}

With this setup, the LLM participates exactly like any other RLCard agent: it receives the full game state, reasons via text prompts, and outputs a legal move.

\begin{figure*}[h]
  \centering
  \includegraphics[width=\linewidth]{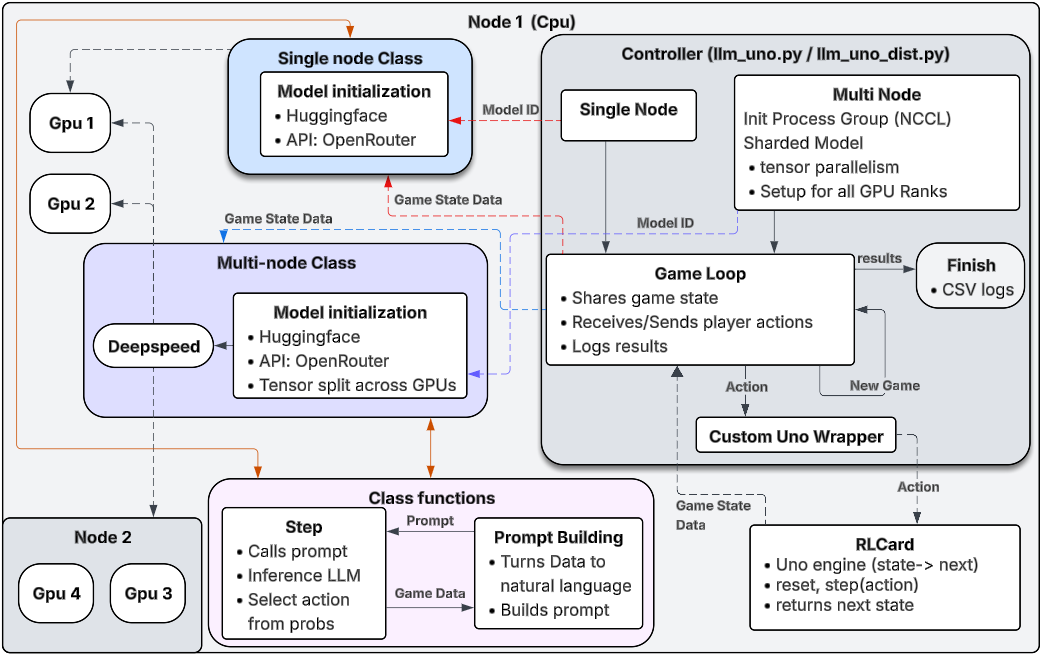} 
  \caption{System architecture for LLM-assisted UNO play in single-node and multi-node modes, showing the controller, agent classes, RLCard integration, and data flow during gameplay.}
  \label{fig:arch}
\end{figure*}

\subsection{System Architecture and Execution}

A single Controller configures the UNO environment, swaps in our custom game class for correct multi-player payoffs, assigns agents, and runs the game loop (see Fig.~\ref{fig:arch}). In the \textit{single-node} setup, we use one of two LLM agent classes—cloze or counterfactual—with the model loaded once from Hugging Face and reused across turns. In the \textit{multi-node} setup, we provide the same two agent classes for distributed inference; we use DeepSpeed to split the model across nodes with tensor parallelism over NCCL. Each turn, the Controller reads the RLCard state and calls the agent’s \texttt{step}: it builds a natural-language prompt from the state and legal actions, runs the LLM, aggregates scores per the chosen prompting method, and returns a legal action. The action is applied in RLCard. Our wrapper applies payoff shaping and logs actions, probabilities, and outcomes to CSV.

\subsection{Statistical Analysis}

To evaluate whether an LLM’s win rate significantly exceeded the baseline, we conducted one‑sided one‑proportion $z$‑tests.  As described by ~\cite{moore2016introduction}, for each model–prompt pairing let \(\hat p\) be the observed win fraction and \(p_0=0.4896\) the baseline probability. We test
\[
  H_0: \hat p = p_0
  \quad\text{versus}\quad
  H_1: \hat p > p_0
\]
using
\[
  z = \frac{\hat p - p_0}{\sqrt{p_0(1-p_0)/n}},
\]
where \(n\) is the number of games (10,000 for most models, 7,000 for LLaMA‑70B). We reject \(H_0\) at \(p < 0.05\).

\begin{figure}[h]
  \centering
    \resizebox{\linewidth}{!}{ 
  \begin{tikzpicture}
    \begin{axis}[
      title={\large LLM Win Rates in 1v1 UNO Matches},
      ybar,
      bar width=10pt,
      symbolic x coords={LLaMA3.2-1B,LLaMA3.1-8B,Mistral-24B,LLaMA3.3-70B},
      xtick=data,
      x=2cm,
      xlabel style={yshift=-5pt},
      ylabel={Win Rate (\%)},
      ymin=46, ymax=54,
      nodes near coords,
      enlarge x limits=0.2,
      x tick label style={
        rotate=20,
        anchor=east,
        xshift=8mm,
        yshift=-4mm
      },
      legend style={at={(rel axis cs:0.5,0.85)},anchor=south,legend columns=3}
    ]
      \addplot+[ybar,bar shift=-12pt] coordinates {
        (LLaMA3.2-1B,51.43)
        (LLaMA3.1-8B,51.22)
        (Mistral-24B,50.62)
        (LLaMA3.3-70B,49.30)
      };
      \addplot+[ybar,bar shift=+12pt] coordinates {
        (LLaMA3.2-1B,51.44)
        (LLaMA3.1-8B,50.03)
        (Mistral-24B,52.10)
        (LLaMA3.3-70B,51.65)
      };

      \path (rel axis cs:0,0) coordinate (xleft);
      \path (rel axis cs:1,0) coordinate (xright);
      \path (axis cs:LLaMA3.2-1B,48.96) coordinate (ybase);
      \draw[black,dashed,thick] (xleft |- ybase) -- (xright |- ybase);

      \addplot+[ybar,bar shift=+12pt, pattern=north east lines,pattern color=black,opacity=0.6,      nodes near coords = {}, forget plot]
          coordinates {
          };
      \addplot+[ybar,bar shift=-12pt, pattern=north east lines,pattern color=black,opacity=0.6,    nodes near coords = {}, forget plot]
        coordinates {
          (LLaMA3.3-70B,49.3)
        };

        \node[rotate=270, anchor=west, font=\footnotesize, yshift=24pt, xshift=10] 
         at (axis cs:LLaMA3.3-70B,51.7) {$p<0.01$};

        \node[rotate=270, anchor=west, font=\footnotesize, yshift=24pt, xshift=10] 
         at (axis cs:LLaMA3.1-8B,49) {$p<0.05$};

        \node[rotate=270, anchor=west, font=\footnotesize, yshift=0pt, xshift=10] 
         at (axis cs:LLaMA3.1-8B,49) {$p<0.01$};
         
        \node[rotate=270, anchor=west, font=\footnotesize, yshift=24pt, xshift=10] 
         at (axis cs:Mistral-24B,51.2) {$p<0.01$};

        \node[rotate=270, anchor=west, font=\footnotesize, yshift=0pt, xshift=10] 
         at (axis cs:Mistral-24B,51.2) {$p<0.01$};
        
        \node[rotate=270, anchor=west, font=\footnotesize, yshift=24pt, xshift=10] 
         at (axis cs:LLaMA3.2-1B,51.7) {$p<0.01$};

        \node[rotate=270, anchor=west, font=\footnotesize, yshift=0pt, xshift=10] 
         at (axis cs:LLaMA3.2-1B,51.7) {$p<0.01$};

      \legend{Counterfactual,Cloze,Statistically Significant Baseline }
    \end{axis}
  \end{tikzpicture}
} 
  \caption{Win rates against a random agent in 1v1 matches using counterfactual and cloze prompting across four LLMs. The dashed line marks the second‑player baseline (48.96\%). Solid bars exceeded the baseline with significance at either $p < 0.05$ or $p<0.001$ as marked. Hatched bars show results that were not statistically significant.}

  \label{fig:llm_1v1_combined}
\end{figure}
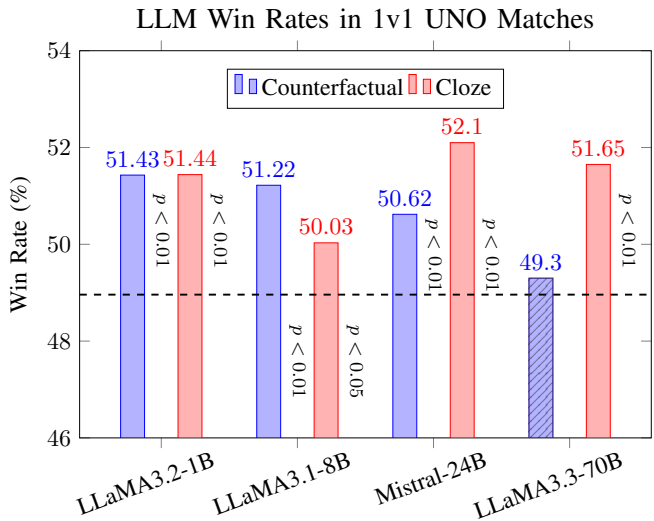

\section{Baseline UNO Win Rate}

Figure~\ref{fig:llm_1v1_combined} reports head‑to‑head win rates of four pretrained LLMs against a random agent under both prompting methods. All four models clear the 48.96\% baseline under at least one prompt, and all but one configuration meet the significance cutoff ($p < 0.05$). Notably, Mistral‑24B achieves the highest overall win rate (52.10\%) with cloze prompting—nearly matching the 4.28\% improvement achieved by RLCard’s DQN agent~\cite{zha2019rlcard}—while even the smallest model LLaMA3.2‑1B surpasses LLaMa3.1-8B. LLaMA3.3‑70B, evaluated on 7,000 games, also improves by 2.74\% under cloze prompting. Despite no fine-tunning, these LLMs performed comparably to RLCard’s DQN agent, highlighting their strong general reasoning capabilities even in a structured decision-making environment such as turn-based card games.

\textbf{Answering RQ1:} Can a pretrained decoder‑only LLM, when playing autonomously, achieve a higher win rate than a random agent in UNO? \textbf{Answer:} Yes-all models exceeded their respective random baselines by statistically meaningful margins with at least one prompting method, including LLaMa3.2-1B also exceeded baseline despite the smaller model size. These results indicate that simple prompting plus inherent pretrained knowledge is sufficient for modest but consistent performance gains over a random policy.

\section{Evaluating LLMs as Agentic Co-op Teammates}

To establish a baseline for cooperative play, we conducted three‑player matches involving two rule‑based agents and a random agent. We chose this “rule–rule–random” configuration because the two deterministic rule‑based players act as stand-ins for traditional human players, while the random agent provides an uninformed co-player, yielding a realistic benchmark against which the aid provided to one of the players by an LLM can be compared. In these trials, seats 0 and 1 achieved win rates of approximately 36.38\% and 35.00\%, respectively, while the random agent won 28.62\%. We therefore adopt the 35.00\% win rate of seat 1 as the unassisted baseline for the assisted position. In these experiments, we ran 10,000 games per model (7,000 for LLaMA3.3‑70B)

\begin{figure}[h]
  \centering
  \resizebox{\linewidth}{!}{
  \begin{tikzpicture}
    \begin{axis}[
      title={\large LLM Supported Seat 1 Win Rates in 3-Player UNO},
      ybar,
      bar width=12pt,
      symbolic x coords={LLaMA3.2-1B,LLaMA3.1-8B,Mistral-24B,LLaMA3.3-70B},
      xtick=data,
      x=2cm,
      xlabel style={yshift=-5pt},
      ylabel={Win Rate (\%)},
      ymin=34, ymax=37,
      nodes near coords,
      every node near coord/.append style={font=\small, yshift=2pt},
      enlarge x limits=0.2,
      x tick label style={
        rotate=20,
        anchor=east, 
        xshift=8mm,
        yshift=-4mm
      },
      legend style={at={(rel axis cs:0.5,0.85)},anchor=south,legend columns=3},
      legend cell align=left
    ]
      
      \addplot+[ybar,bar shift=-12pt] coordinates {
        (LLaMA3.2-1B,34.58)
        (LLaMA3.1-8B,35.49)
        (Mistral-24B,35.74)
        (LLaMA3.3-70B,35.67)
      };
      \addplot+[ybar,bar shift=+12pt] coordinates {
        (LLaMA3.2-1B,35.76)
        (LLaMA3.1-8B,35.01)
        (Mistral-24B,35.26)
        (LLaMA3.3-70B,35.96)
      };
    
      \path (rel axis cs:0,0) coordinate (xleft);
      \path (rel axis cs:1,0) coordinate (xright);
      \path (axis cs:LLaMA3.2-1B,34.91) coordinate (ybase);
      \draw[black,dashed,thick] (xleft |- ybase) -- (xright |- ybase);

      \addplot+[ybar,bar shift=+12pt,pattern=north east lines,pattern color=black,opacity=0.6,     nodes near coords = {}, forget plot]
        coordinates {
            (LLaMA3.2-1B,35.76)
            (LLaMA3.1-8B,35.01)
            (Mistral-24B,35.26)
        };
      \addplot+[ybar,bar shift=-12pt, pattern=north east lines, pattern color=black,opacity=0.6,   nodes near coords = {}, forget plot]
        coordinates {
            (LLaMA3.2-1B,34.58)
            (LLaMA3.1-8B,35.49)
            (Mistral-24B,35.74)
            (LLaMA3.3-70B,35.67)
        };
        \node[rotate=270, anchor=west, font=\footnotesize, yshift=24pt, xshift=10] 
         at (axis cs:LLaMA3.3-70B,35.96) {$p<0.05$};

      \legend{Counterfactual, Cloze}
    \end{axis}
  \end{tikzpicture}
}
  \caption{In a three‑player “1v1v1” setting with two rule‑based agents in seats 0 and 1 and an LLM in seat 2, we compare the Rule 1 seat win rates across four LLMs under cloze (red) and counterfactual (blue) prompting (10,000 games per model and prompt; LLaMA3.3‑70B on 7,000 games). The dashed line marks the unassisted baseline. Solid bars exceeded the baseline with significance $p < 0.05$. Hatched bars show results that were not statistically significant.}

  \label{fig:rule1_combined}
\end{figure}
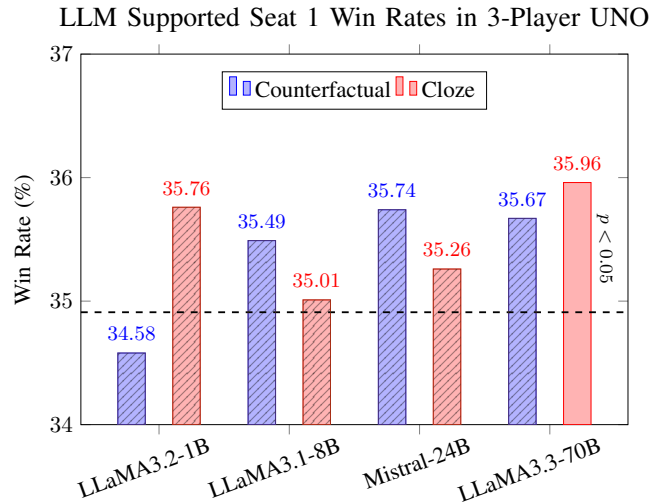

In all assistance trials, the LLM occupied seat 2, assisting the rule‑based agent in seat 1. Under cloze prompting (Figure~\ref{fig:rule1_combined}), LLaMA3.2‑1B (35.76\%) approached and LLaMA3.3‑70B (35.96\%) surpassed the significance threshold, $p < 0.05$. Under counterfactual prompting, Mistral‑24B (35.74\%) similarly yielded a near-significant result. LLaMA3.3‑70B with cloze prompting delivered the largest assistance gain, demonstrating that prompt design and model scale interact nonlinearly in cooperative settings.

\textbf{Answering RQ2.} Can a pretrained LLM, when deployed as a cooperative partner, improve a teammate’s win rate? We compared each model–prompt configuration’s assisted win rate to the 35.00\% (100,000 games) unassisted baseline using one‑sided one‑proportion $z$‑tests. Three configurations approached or achieved statistical significance: LLaMA3.2‑1B with \emph{cloze} prompting, Mistral‑24B with \emph{counterfactual} prompting, and LLaMA3.3‑70B with \emph{cloze} prompting, which was the only configuration with $p < 0.05$. These results demonstrate that prompting methods can influence cooperative gains, with LLaMA3.3‑70B under cloze prompting yielding the largest observed improvement.

\textbf{Answering RQ3.} To what extent does scaling the LLM’s size impact its autonomous and cooperative effectiveness? In solo play, mid‑sized Mistral‑24B nearly matched LLaMA3.3‑70B’s top cloze win rate (52.10\% vs.\ 51.65\%), and even the smallest LLaMA3.2‑1B outperformed LLaMA3.1‑8B (51.44\% vs.\ 50.03\%). In cooperative trials, LLaMA3.3‑70B was the only model–prompt combination to surpass the significance threshold, though LLaMA3.2‑1B and Mistral‑24B produced near‑significant results. These findings suggest that while scale may help, it is not strictly conclusive—LLaMA3.2‑1B rivaled or exceeded larger models in several cases, indicating that architecture and prompting may outweigh raw size in certain settings.

\textbf{Answering RQ4.} How do prompting techniques—cloze versus counterfactual—affect decision quality? We found prompt efficacy to be model‑dependent. In autonomous play, cloze prompting yielded the highest win rates for LLaMA‑1B, Mistral‑24B, and LLaMA‑70B. In cooperative trials, three model–prompt combinations approached or surpassed the significance threshold: LLaMA‑1B with cloze prompting, Mistral‑24B with counterfactual prompting, and LLaMA‑70B with cloze prompting. Overall, cloze prompting yielded more consistent gains across models, suggesting it may be the more robust strategy in both solo and collaborative settings.

\section{Conclusion}

This study evaluated four decoder‑only LLMs integrated into the RLCard UNO framework under zero‑shot conditions. In 1v1 matches, every model outperformed the second‑player baseline under at least one prompting strategy, with cloze prompting driving the strongest results (e.g., Mistral‑24B reaching 52.10\%). In three‑player cooperative trials, three model–prompt combinations—LLaMA3.2‑1B with cloze, Mistral‑24B with counterfactual, and LLaMA3.3‑70B with cloze—produced reliable assistance gains, with only LLaMA3.3‑70B achieving statistical significance ($p < 0.05$). These findings demonstrate that pretrained LLMs can not only reason effectively on their own but also deliver statistically significant cooperative gains, although results vary by model scale and prompting method.

\section{Limitations}

This work has several limitations. First, UNO’s cooperative mechanics are inherently constrained: support arises mainly through action cards (Reverse, Skip, Draw Two, Wild Draw Four), and the strategic impact of each card depends heavily on seating order and hand composition. Second, we evaluated cooperation exclusively alongside fixed rule‑based agents; learning‑based agents (e.g., DQN) or human partners might interact differently with an LLM teammate and unlock additional gains. Third, our prompt engineering was limited—we relied on raw prompt text with manually added tags rather than systematic templating frameworks (e.g., Hugging Face prompt templates), which may have introduced inconsistencies. Fourth, the largest model (LLaMA3.3‑70B) was only evaluated on 7,000 games per configuration instead of the 10,000 used for the other models, reducing statistical power for borderline cooperative effects. Finally, we did not assess larger or proprietary models (e.g., GPT‑4) due to resource constraints; such models might yield more substantial cooperative improvements.

\section{Future Work}

Future work should continue to explore the potential of agentic LLMs as assitants that provide aid to accomplish the strategic objectives of other individuals. Specifically, work should focus on: 

\begin{itemize}
  \item \textbf{Equalize trial counts:} Run 10{,}000 games for all models—including the largest ones—to improve statistical power and directly compare across scales.
  \item \textbf{Alternative cooperative environments:} Evaluate LLMs in more complex games to test collaboration under these stricter rules and interactions.
  \item \textbf{Expanded model list:} Include larger models (e.g., GPT‑4, Claude) to examine how parameter scale impact decision quality in solo and team scenarios.
  \item \textbf{Varied teammate dynamics:} Pair LLMs with adaptive agents (e.g., DQN) and human players to assess how different partner types influence cooperative success.
\end{itemize}

\bibliographystyle{IEEEtran}
\bibliography{text_ref}

\end{document}